# A HYBRID RULE BASED FUZZY-NEURAL EXPERT SYSTEM FOR PASSIVE NETWORK MONITORING

**AZRUDDIN AHMAD, GOBITHASAN RUDRUSAMY,**
**RAHMAT BUDIARTO, AZMAN SAMSUDIN, SURESRAWAN RAMADASS.**

Network Research Group
School of Computer Science,
Univ. Science Malaysia -11800 Minden, Pulau Pinang
MALAYSIA
Tel.: +60-4-659-4757
Fax: +60-4-657-3335

E-mail: azru@nrg.cs.usm.my, gobithasan@nrg.cs.usm.my,
{rahmat,azman,sures}@cs.usm.my

# A HYBRID RULE BASED FUZZY-NEURAL EXPERT SYSTEM FOR PASSIVE NETWORK MONITORING

## ABSTRACT
An enhanced approach for network monitoring is to create a network monitoring tool that has artificial intelligence characteristics. There are a number of approaches available. One such approach is by the use of a combination of rule based, fuzzy logic and neural networks to create a hybrid ANFIS system. Such system will have a dual knowledge database approach. One containing membership function values to compare to and do deductive reasoning and another database with rules deductively formulated by an expert (a network administrator). The knowledge database will be updated continuously with newly acquired patterns. In short, the system will be composed of 2 parts, learning from data sets and fine-tuning the knowledge-base using neural network and the use of fuzzy logic in making decision based on the rules and membership functions inside the knowledge base. This paper will discuss the idea, steps and issues involved in creating such a system.

## INTRODUCTION
The dependency on network administration has risen due to an increasing amount of mission critical applications dependant on the network. The task of network administration requires applied repetitive task of recognition and evaluation of conditions which are among the main applications of Artificial Intelligence [5]. This makes network monitoring a potential user of Artificial Intelligence.

Current network monitoring systems are non adaptable rule based systems. Such systems are NETmon from NRG (Network Research Group) at USM [4]. Existing network monitoring systems with artificial intelligence are scarce or nonexistent.

An improved network monitoring system demands the need for an automated diagnosis method, a simulation of an expert (network administrator) knowledge, and decision making process. This will require the use of artificial intelligence methods. This system uses the hybrid approach to overcome disadvantages of using only one single method [5,7,8]. Rather than performing the whole task with one technique that is not ideal for all aspects, a couple of techniques are used as appropriates [2]. Here, we will discuss a hybrid system consisting of rules derived from knowledge of an expert, fuzzy logic [7] to diagnose conditions and artificial neural network for refining the membership functions to make the system adaptable.

Currently, most systems employing AI method in network monitoring are expert or rule systems for fault diagnosis [2]. Other proposed methods including AI methods are detection from statistical deviations from regular observed behavior monitoring [9], threshold in a time series model, adaptive monitoring systems, clustering methods and markov models [3].

Performance Monitoring, Fault Diagnosis and Network Control and Diagnosis are three functions associated with AI techniques [1]. The processes involved forms a natural hierarchy:
- low level – anomaly detection
- diagnosis level – find cause
- high level – identify root cause, devise course of action

There are two types of approach that could be taken to tackle network monitoring and diagnosis which are monitoring and diagnosis or event correlation approach.

Considerations put into and some of the issues involved to create an expert system for network monitoring are:
- Different network conditions on each network. Although the same basic rules will still apply.
- Defining predefined rules can cause redundancy and inadaptability of the system whereas a system without any rules might not produce correct decisions. The need of using rules has to be decided on the type of system planned.
- The use of linguistic terms such as *many, few, probably* in rules for variables to allow a diagnosis process close to natural human expression.
- The information needed for the monitoring, learning and decision process requires a high volume of daily data to process.

The proposed system is on the *low detection level* monitoring and diagnosis with adaptive capabilities to adjust to different network conditions. We will use the real time monitoring and diagnosis approach to prevent the problem of high amount of data to be processed. The data will be processed when it arrives which will also allow for real time monitoring. The issues above will be solved by creating an adaptable rule based system using neural network. Trustworthiness of the system can be obtained by the predefined rules and neural network will make the system adaptable. Fuzzy logic is used for making weighted decisions and minimize. This will create a more "fuzzy" [8] decision and minimize the number of false alarms. The decision process is in real time.

## 2 DESIGN

The core of the system is the classic approach, rules based on formulation of expert rules. A dual knowledge base system is needed. One to hold the rules and the other for the membership values by the rules. Fig.1 shows the architecture of the system.

Two modes of processing will be run concurrently: A learning mode to adjust and fine tune the membership values of the rules and an execution mode for diagnosing the network conditions. The other will be the monitoring and diagnosis mode to look for abnormalities and output an assessment.

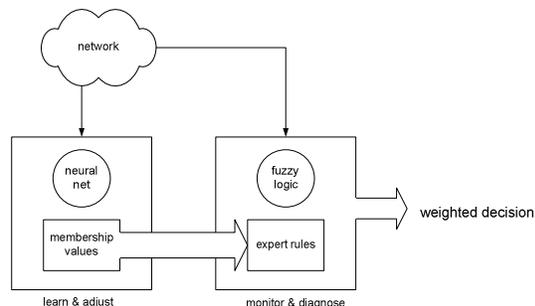

Fig 1: Architecture of the system

At first deployment, the membership values will be nonexistent. The value will be obtained following an algorithm based on the rules for the particular membership function (see Fig.2). Further values are obtained at timely intervals. Adjustment is done to the membership value based on the perceptron learning process [8]. The membership values are stored based on the time and adjustment is also done based on this. The comparison based on the time variable is important to make out patterns from the data available. Decisions made on anomalous traffic behavior will be dependant on the values of the membership functions.

**Rules**
Abnormal traffic behavior could be classed as outages, configuration changes, flash crowd and abuse. Recognizing and identifying such behaviors are often based on ad hoc methods developed from years of experience in managing networks [3]. Basic patterns of such behavior are known and simple rules based on known conditions are devised by experts. Researches into patterns of network activities are also being done to create such rules. These rules serve as the basic knowledge of the system and are basic enough to apply to any network. To apply to this system, fuzzy rules are used. The fuzzy rules will be based on domains created.

**Database**
The system consists of dual database. To create a proper expert system, a database containing predefined basic rules which act as a knowledge base is needed. The second database will contain learned values to apply to the rules. These values are learned, adjusted and fine tuned by using the *perceptron learning* approach. At the beginning of deployment of the system to an unknown network, the database is empty. The values are learned and always adjusted. This will make the values ever changing to conditions and are specific to the certain network.

**Creating & refining membership function**
Membership functions are used in learning distinctive characteristics of the network to be monitored. The membership functions are based on the predefined expert rules. The rules require that certain establishment of domains where a condition will be grouped into. The membership values of these domains are learned and adjusted to suite the condition of the network and to allow for new and changing data. This will make the rules more robust and can easily evolve to new network conditions.
The refinement process follows the neural net approach of learning and adaptation. The algorithm for this process is as follows:

For each domain there are a predefined number of membership functions (see Fig.2). The value for the transition from one membership function will be obtained using these steps.

1. The average data value of the domain over a predefined time period is obtained.
2. The average data value is calculated. This value will divide the data into two regions.
3. For every region, step 2 is repeated. This step will divide the data set into 3 regions.
4. Step 2 and 3 might be recursively repeated to create the desired amount of membership functions.

For each membership function value, *perceptron learning* will be used to update and train the membership value weights in the knowledge base.

| |
|---|
| *Abnormal network use* |
| Domain & membership function: network utilization (low, normal, extreme) <br> bandwidth use (low bandwidth use, normal, high) |
| Rules: traffic extreme, at usually low use then network condition abnormal <br>   traffic extreme, at usually high use then network condition normal <br>   traffic low, at usually low bandwidth then normal <br>   traffic extreme, at usually low bandwidth then network condition abnormal <br>   traffic normal, at usually normal network bandwidth then network condition normal. |

Fig.2 – Expert Rule Example*

*The membership function values for these conditions are learned by the system (See Fig.3).
*The rules are based on fuzzy rule format. Decisions are done by using fuzzy logic. Fig. 4 depicts example of fuzzy decision process in the system.

Perceptron learning:
  $w_i(p+1) = w_i(p) + \Delta w_i(p)$
  where $\Delta w_i(p) = \alpha * x_i(p) * e(p)$
  $\alpha$ = learning rate
  $e(p) = y_d(p) - y(p)$
  $y_d(p)$ = desired output (previous weight)
  $y(p)$ = actual output

  For $w(1)$ = first value obtained
  $w(2) = w(1) + \Delta w_i(1)$
  .
The steps above are repeated continually until it stabilizes at a certain value for the membership function.

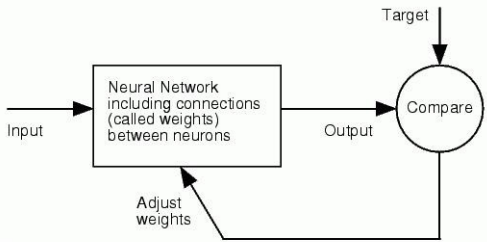

Fig 3: learning process

**Decision process**
The decision process uses fuzzy logic to come out with a weighted decision and reduce the numbers of false alarms [3]. Fuzzy logic is very good at handling probability and uncertainty. This will allow a human like decision to be made. It will also overcome the threshold border decision problem [3] in current systems. The process will be based on the weights of the membership function of the inputs which will also result in a weighted decision (Fig 4). Based on the significance of the condition, the proper actions can be taken. The decision derived will be as similar to human decision as possible.

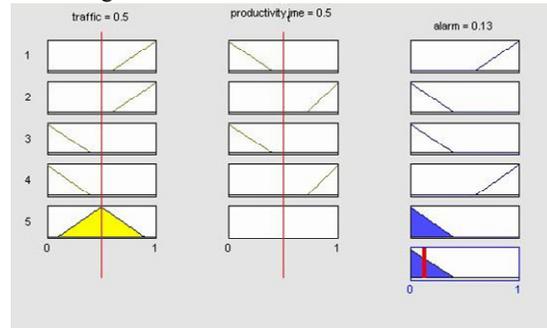

Fig 4.1 condition normal at value 0.13

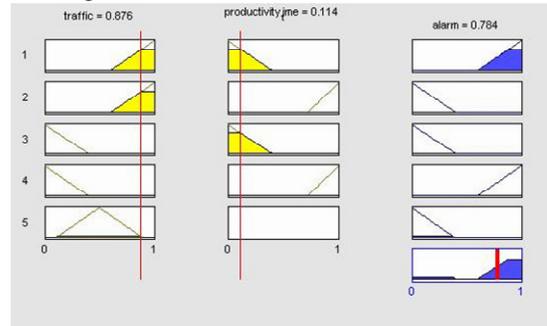

Fig 4.2 condition abnormal at value 0.784

Fig.4 Examples of fuzzy decision process

**Classing of condition based on learned knowledge**
Some inputs of the rules are conditions that have to be decided and learned. An example of this (Fig.2) is the bandwidth domain. The condition has to be learned first and than applied to for the value to be obtained.

**Anomaly recognition based on membership function**
With the learning, updating and fine tuning process described here, an intelligent system that simulates a network administrators decision and learning process with the ability to adapt, learn and give out weighted decision can be created.

## 3 IMPLEMENTATION – iNETMon

Implementation of the system is currently in progress. The intelligent module is being formed for the network monitoring software NETmon, a passive network monitoring software created by Network Research Group, University Science of Malaysia [4].

NETmon is a passive real-time network monitoring software with tools to passively monitor the network. It is based on the Windows platform environment. The current version does not have artificial intelligence.

The data for the module will be obtained from the NETmon interface. The iNETMon module is for warning and alarm creation. There is an alarm module in the current version but it is threshold based. The drawback of this approach is that it is non adaptable, dependant on the network administrator to set and change the threshold values. There are no weights to the seriousness of the alarms created. Also it will be subjected to the threshold problem [3]. The iNETMon module will overcome the drawback of his approach.

With the ANFIS for NETmon approach the alarms will have values to it. With this, the system can decide the action to be taken in relation to the seriousness of the alarm received. For instance a *probable* network misuse will only be logged but a *sure* network breach decision can be added with further action of sending an *sms* to the network administrator to inform him immediately. The system could also take further action by taking proactive measures such as taking down the breached server. The use of artificial intelligence in NETmon promises an unsupervised and better network administration.

## 4 PRELIMINARY RESULTS

Patterns inside of daily network traffic are the main dependency of the system. Normal human observations can make out such patterns. Still, unless proper research into such patterns are made, further knowledge about the behavior of these patterns could not

The data discussed here are based on the observation of network traffic at Computer Science School, University Science of Malaysia (Fig 5). The data obtained shows promising prospect in finding data patterns inside network traffic data. Research is currently being done on obtainable network patterns and conditions. These will be used to create rules needed by the proposed system. Similar data on network patterns could be found in Barford & Plonka [3].

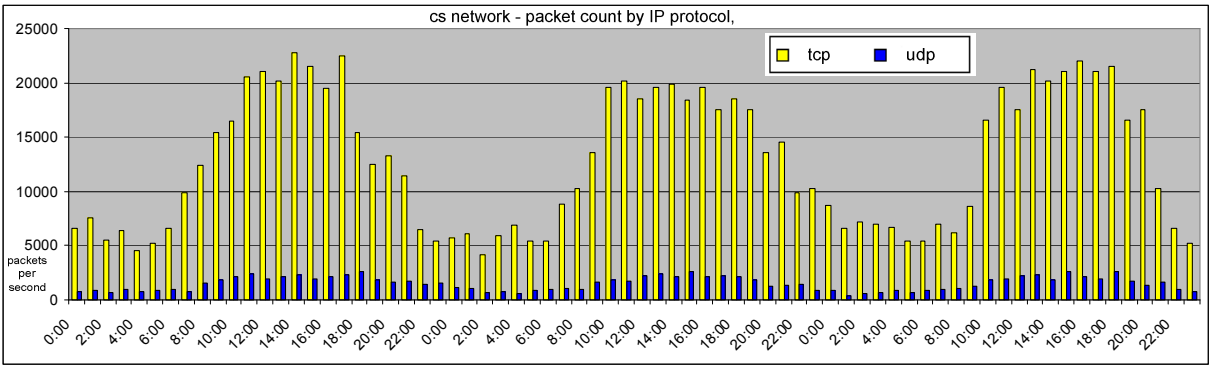

Fig. 5 : packet count per second in a typical 72 hour period at NRG, USM.

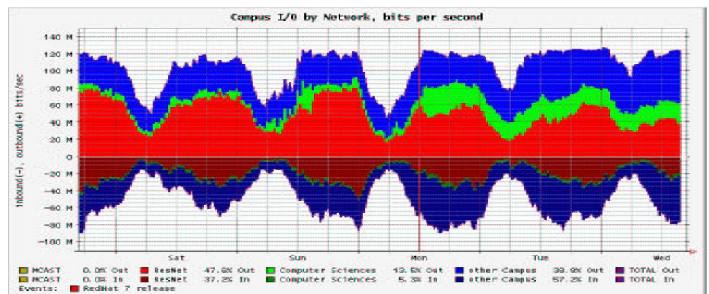

Fig. 6: example of flash crowd behavior [3]

## 5 DISCUSSION

The proposed system is still dependant on predefined rules. This is because of the need for the system to know certain patterns of network problems. Such problems are only known by experts of the field and could not be learned by the system without some sort of supervision. Still, most of the problems discussed above are caused by the use of the ES (Expert System) rules. A better system would be a system that learns rules by itself inductively. Such similar systems are proposed in Hermann [5]. Further work should be done on this.

The learning curve of the proposed system is yet to be seen. Also there are some other issues that arise on the configuration and performance of the network. The current netmon also has some issues in network configurations whereas configurations using routers and switches experience difficulties to be monitored.

The need for high amounts of processing to achieve real time processing is also a key factor. The sheer amount of data and rules to filter through in high speed networks might cause the aim for real time monitoring unobtainable.

The dependency on the rules is the main drawback of this system. Another approach is the system to not have rules altogether. Such systems will only be dependant on patterns. Fuzzy clustering is one of the promising approaches available for this purpose.

## 6 CONCLUSION

With the proposed system of incorporating different AI approaches, it will overcome the limitations of a system using only one or two approaches [2]. The limitations overcame:
- Ability to handle new and changing data. Robust when faced with unforeseen situations
- Learning and fine tuning from conditions encountered
- Able to handle probability and uncertainty
- Adaptable if the network evolves.

Still there are problems unsolved and new problems created by using the method explained:
- Past experiences are not stored. Past success or failures are not remembered or applied to current problems. The experience is just in a value only.
- Rule base system will require maintenance at some time to add rules or delete obsolete rules.
- The domain must be well understood and thought out. This is not entirely possible in domains such as fault management.

## 7 FUTURE WORKS

At the time of writing we are in the process of gathering and archiving data and researching on anomalies based on traffic on a part of network in our lab. Additional data will also be gathered from other networks to do comparison. We are also at the early stages of applying various statistical techniques to analyze the data and creating rules.

The rules will then be tested on the created modules to establish if the intended results are obtained.

We are also researching other approaches available to create a truly intelligent system. One promising step is by creating a system not dependent on rules, but instead uses cluster patterns in making decisions.